\documentclass[conference]{IEEEtran}
\IEEEoverridecommandlockouts

\usepackage{cite}
\usepackage{amsmath,amssymb,amsfonts}
\usepackage{mathtools}
\usepackage{algorithmic}
\usepackage{graphicx}
\usepackage{textcomp}
\usepackage{xcolor}
\usepackage{url}
\usepackage{xurl}
\usepackage{bm}
\usepackage{subfigure}
\usepackage{placeins}
\usepackage{booktabs}
\usepackage{comment}
\usepackage{makecell}
\usepackage{float}
\usepackage[hidelinks]{hyperref}
\usepackage{threeparttable}
\usepackage{soul}

\def\BibTeX{{\rm B\kern-.05em{\sc i\kern-.025em b}\kern-.08em
    T\kern-.1667em\lower.7ex\hbox{E}\kern-.125emX}}
\begin{document}

\title{Reservoir Computing inspired  \\ Matrix Multiplication-free Language Model
}

\author{\IEEEauthorblockN{Takumi Shiratsuchi${}^\text{*}$, Yuichiro Tanaka${}^\text{*\dag}$, and Hakaru Tamukoh${}^\text{*\dag}$}
\IEEEauthorblockA{\textit{*Graduate School of Life Science and Systems Engineering, Kyushu Institute of Technology, Japan} \\
\textit{\dag Research Center for Neuromorphic AI Hardware, Kyushu Institute of Technology, Japan}\\
}
}

\maketitle

\begin{abstract}
Large language models (LLMs) have achieved state-of-the-art performance in natural language processing; however, their high computational cost remains a major bottleneck. In this study, we target computational efficiency by focusing on a matrix multiplication free language model (MatMul-free LM) and further reducing the training cost through an architecture inspired by reservoir computing. 
Specifically, we partially fix and share the weights of selected layers in the MatMul-free LM and insert reservoir layers to obtain rich dynamic representations without additional training overhead. Additionally, several operations are combined to reduce memory accesses. Experimental results show that the proposed architecture reduces the number of parameters by up to 19\%, training time by 9.9\%, and inference time by 8.0\%, while maintaining comparable performance to the baseline model.

\end{abstract}


\section{INTRODUCTION}
Large language models (LLMs), for example, Llama3 \cite{llama3etal} and DeepSeek-V3 \cite{deepseekv3etal}, achieve performance comparable to humans in various tasks such as sentence generation and dialogue\cite{Khraisha_2024}. Consequently, companies and public institutions are rapidly deploying them in real-world products and services.
Researchers have also demonstrated strong results when they apply LLMs to legal case retrieval \cite{legalRag, padiu2024extent}, code generation with automatic debugging \cite{code_llm}, medical question-answering systems \cite{medicineLLMsurvey, medicalLLM}, and robot action planning \cite{LLM-for-robot-survey, robot_llm, Yamao_ICRA2024, Yano_ROMAN24}.

Despite these advances, the computational cost of both training and inference remains prohibitively high for most computing environments. For example, training a practical large-scale model, Llama3 405 billion (B) \cite{llama3etal}, requires approximately 50 days even when using 16,000 NVIDIA H100 graphics processing units (GPUs), each equipped with 80 GB of memory. Inference is also computationally expensive: generating 256 tokens with a context length of 4,096 tokens using 16 H100 GPUs requires approximately one second of computation.

A primary factor contributing to the high computational cost of LLM training and inference is the large memory footprint arising from the massive number of parameters and their bit widths. For instance, inference of Llama3 405 B without quantization requires approximately 750 GB of memory solely for storing model parameters. To mitigate this issue, previous studies have proposed methods to reduce the computational and memory costs of LLMs, including quantization of pre-trained models \cite{LLMint8}, \cite{LLMint4} and low-rank decomposition \cite{LoRD}. However, these approaches provide only limited reductions in parameter count and memory usage.

This study focuses on the matrix multiplication-free language model (MatMul-free LM) \cite{MM-freeLM} to address the memory usage problem. MatMul-free LM is a type of LLM where the weights are quantized to ternary values \{+1, 0, -1\} through quantization-aware training (QAT) \cite{QandT}, \cite{LSQ+}, \cite{QATTransformer}, leading to a significant reduction in memory usage \cite{MM-freeLM}. Additionally, MatMul-free LM replaces attention mechanism used in most LLMs with a matrix multiplication-free linear gated recurrent unit (MLGRU). This change contributes to a further decrease in computational costs.
Although MatMul-free LM reduces the bit width of individual weights, the problem of its large number of parameters remains, limiting its effectiveness.
In addition, pre-training all parameters in the model incurs substantial computational costs.

To solve these issues, this study introduces the concept of reservoir computing (RC) \cite{ESN}. RC is a computationally efficient recurrent neural network (RNN) where both the input and reservoir layers have fixed weights, while only the output layer is trainable \cite{ESN}. RC is known to perform well despite the small number of trainable parameters\cite{BPforRC}. Furthermore, research on efficient implementation methods using dedicated hardware and physical dynamics has advanced considerably, and is expected to yield practical developments\cite{Usami_MaterialRC, LUTNetRC_FPT23}.

Here, we propose an RC inspired MatMul-free LM (RC MatMul-free LM), applying RC to MLGRU in MatMul-free LM.
This research is the first attempt to introduce an RC-extended component into LLM.
The proposed RC MatMul-free LM reduces not only the bit width of each parameter but also the total number of parameters by fixing and sharing parameters, based on the findings reported in previous studies\cite{albert, reservoir-transformer}, which show that performance degradation remains small even when parameters are shared across layers.

\section{RELATED WORK}
\subsection{Architecture of LLM}
\begin{figure}[!t]
    \centering
    \subfigure{\includegraphics[width=0.2\textwidth]{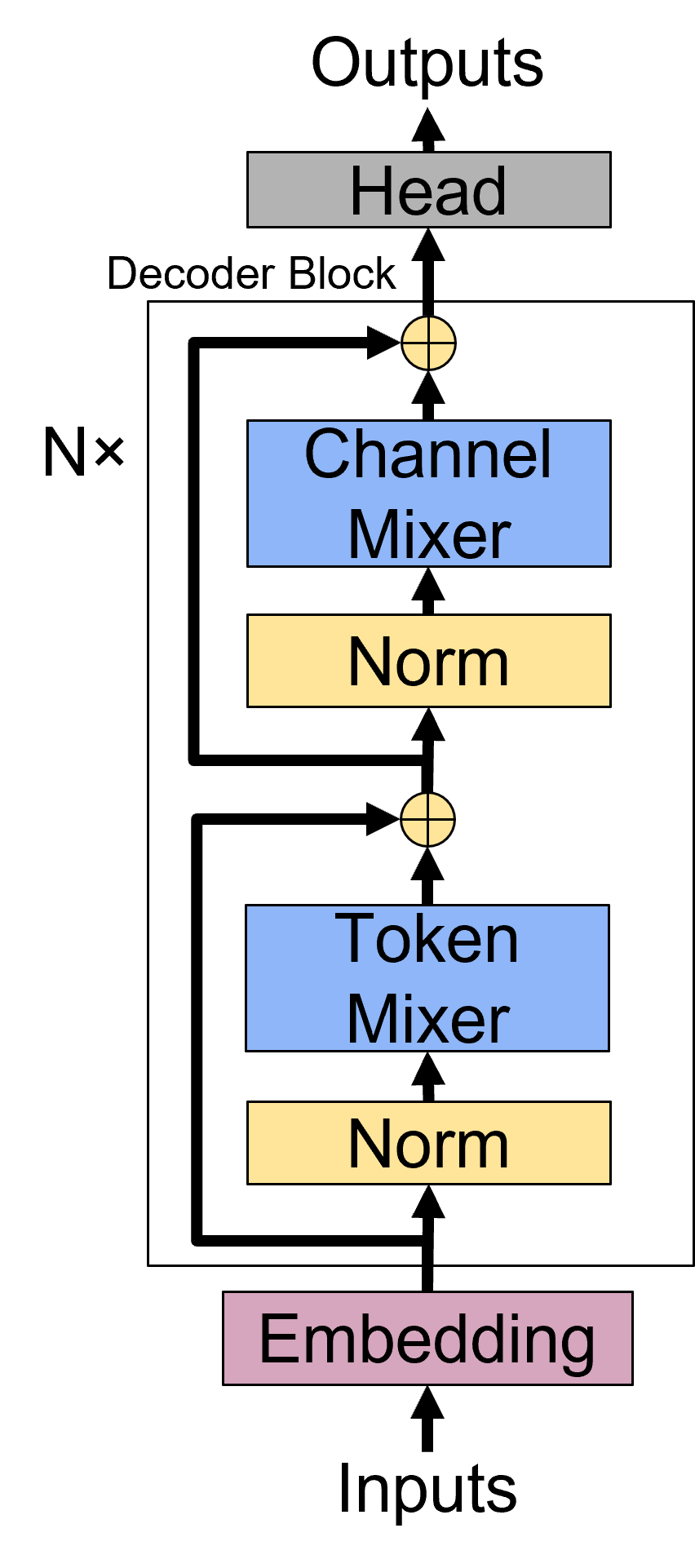}}
    \subfigure{\includegraphics[width=0.2\textwidth]{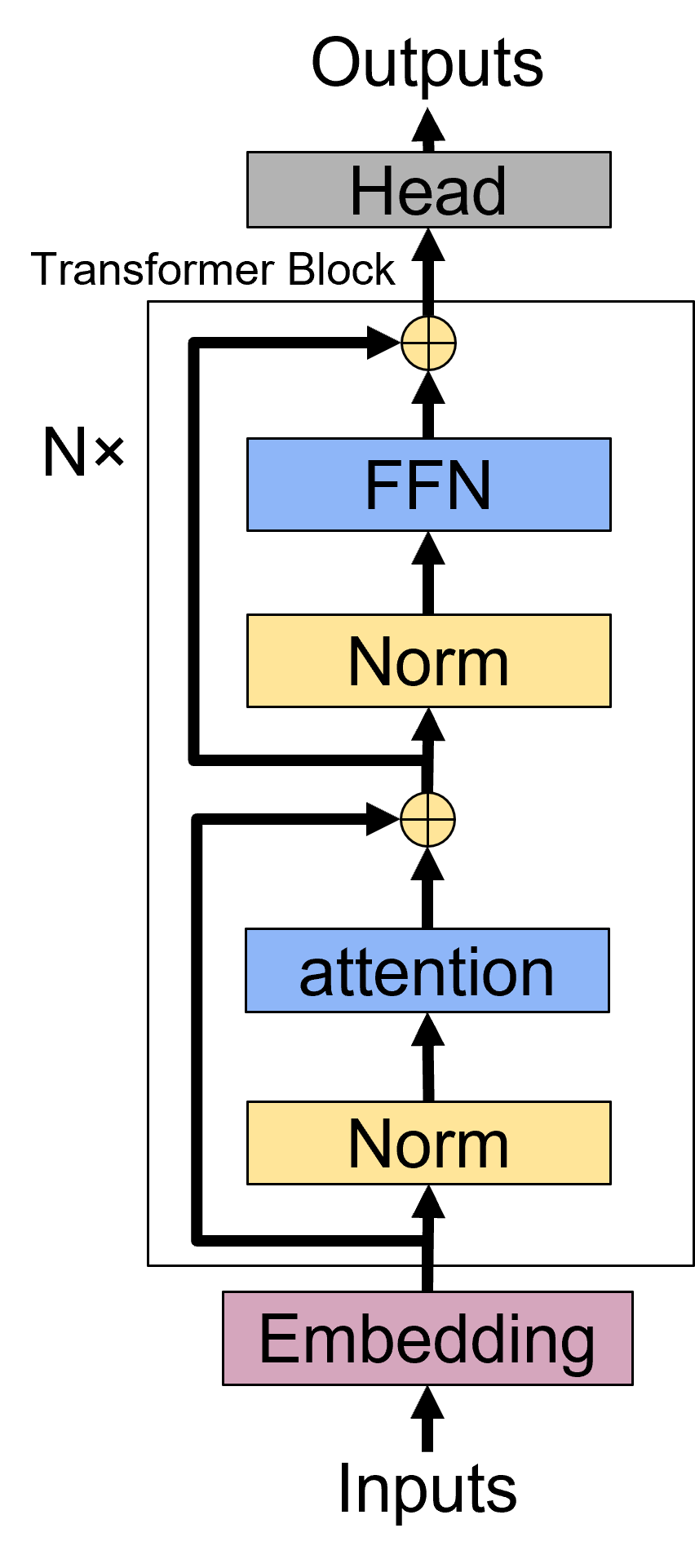}}
\caption{Left : architecture of general LLM right : architecture of transformer-based LLM}
\label{LLM component}
\end{figure}

Fig. \ref{LLM component} shows an architecture of generic LLM and transformer-based LLM \cite{attention-is-all-you-need}.
As shown on the left side of Fig. \ref{LLM component}, LLM mainly consists of six components: embedding layer, normalization (Norm) layer, token-mixer, channel-mixer, head, and residual connections that directly add the inputs \cite{metaformer}. The token-mixer and channel-mixer account for the majority of parameters.
Almost all LLMs employ a feed-forward network (FFN) as the channel-mixer.
In contrast, there are various architectures for the token-mixer. LLMs, with the most widely used architecture, such as Llama3 \cite{llama3etal}, adopt the transformer \cite{attention-is-all-you-need} architecture and use attention mechanism as the token-mixer, as shown on the right side of Fig. \ref{LLM component}.
Collectively, the component consisting of the FFN, attention, Norm, and residual connection is referred to as the transformer block.
Despite its prevalence, the attention mechanism suffers from the issue that the computational cost scales quadratically with the input sequence length. To address this problem, alternatives have been proposed, such as Mamba \cite{Mamba}, which employs a state space model instead of the attention mechanism; Receptance Weighted Key Value (RWKV) \cite{peng-etal-2023-rwkv}, which replaces attention mechanism with an RNN; and a hierarchically gated linear RNN (HGRN) \cite{hgrn}.

HGRN reproduces the attention mechanism by RNN. While simple RNNs cannot adequately capture the long-term dependencies that the attention mechanism can, HGRN employs a hierarchically designed forget gate to capture both short and long-term dependencies. Specifically, the lower bound of the forget gate is trained through the cumax activation function \cite{ON-LSTM} and is constrained to increase monotonically with depth: layers nearer the embeddings receive a small lower bound, while layers nearer the output head receive a large one. Consequently, the lower layers whose forget gates take small values retain past states only briefly and capture short-term dependencies, whereas the upper layers whose gates take larger values retain past states for much longer and capture long-term dependencies \cite{hgrn}.

\subsection{Quantization and Other Model Compression for LLMs}
In LLMs, a scaling law \cite{scaling_raw} indicates that improving performance requires increasing model parameters, training data, and computational resources, thereby necessitating greater memory usage and longer processing times for both training and inference.
To address the increasing memory usage of inference associated with performance, quantization of trained LLMs has been widely considered. Specifically, quantization of most weights to int8 \cite{LLMint8} or 
int4 \cite{LLMint4}, and quantization using QLoRA \cite{qlora} to 2-bit or 3-bit NormalFloat data types \cite{apiq, ra-lora, ai-qlora}, where each value represents a brain floating-point (bfloat) 16 \cite{google_bfloat} value, have been investigated. Furthermore, OPTQ \cite{optq} quantizes the weights to 3-bit per layer and utilizes a kernel optimized for 3-bit quantized weights to minimize memory accesses and achieve fast inference. However, there are limitations to these methods, such as the 
trade-off between quantization error and memory usage, and the need for float16 or bfloat16 operations to maintain accuracy, leading to increased computation time in inference.

QAT \cite{QandT, LSQ+, QATTransformer} is effective for constructing high-performance binary or ternary LLMs. 
As research into developing high-performance binary and ternary LLMs using QAT, Wang et al. showed that when a 1-bit model is scaled, the performance of the model with quantization approaches that of the model without quantization \cite{wang2023bitnet}.
Ma et al. showed that BitNet with 3.9B parameters quantized to three values performs better than Llama \cite{llama, llama2} with 3B parameters without quantization, using more memory for inference \cite{ma2024era}. However, the training process requires the use of float16 and other data types, and the computational cost is enormous, especially for pre-training.

There are also methods for accelerating LLMs and reducing their memory usage by fixing and sharing parameters.
Reservoir transformer \cite{reservoir-transformer} has shown that the parameters of some transformer blocks can be fixed without training and shared by transformer blocks in different layers, thereby reducing the model size. Instead of fixed transformer blocks, an architecture using bi-directional GRU (BiGRU) \cite{gru} with all weights fixed, called BiGRU reservoir, is also proposed. However, it only introduces an untrained BiGRU immediately after the transformer block, and no consistent performance improvement has been observed. A lite bidirectional encoder representations from transformers (ALBERT) \cite{albert} showed that sharing the attention mechanism parameters in each layer caused little performance degradation. However, no studies have confirmed this property for models combining RNNs and transformers, including RWKV.

Other methods for reducing parameters include approximating the weight matrix of a trained LLM with multiple low-dimensional matrices, called low-rank decomposition \cite{LoRD}, and reducing weights of low importance, called pruning\cite{Pruning}, but their effectiveness is limited.

\subsection{MatMul-freeLM}
MatMul-free LM \cite{MM-freeLM} is one of the extremely lightweight LLMs that enables the implementation of matrix multiplication using logical operations by quantizing the parameters to ternary values \{+1, 0, -1\}. 

\begin{figure}[!t]
    \centerline{\includegraphics[width=1.0\linewidth]{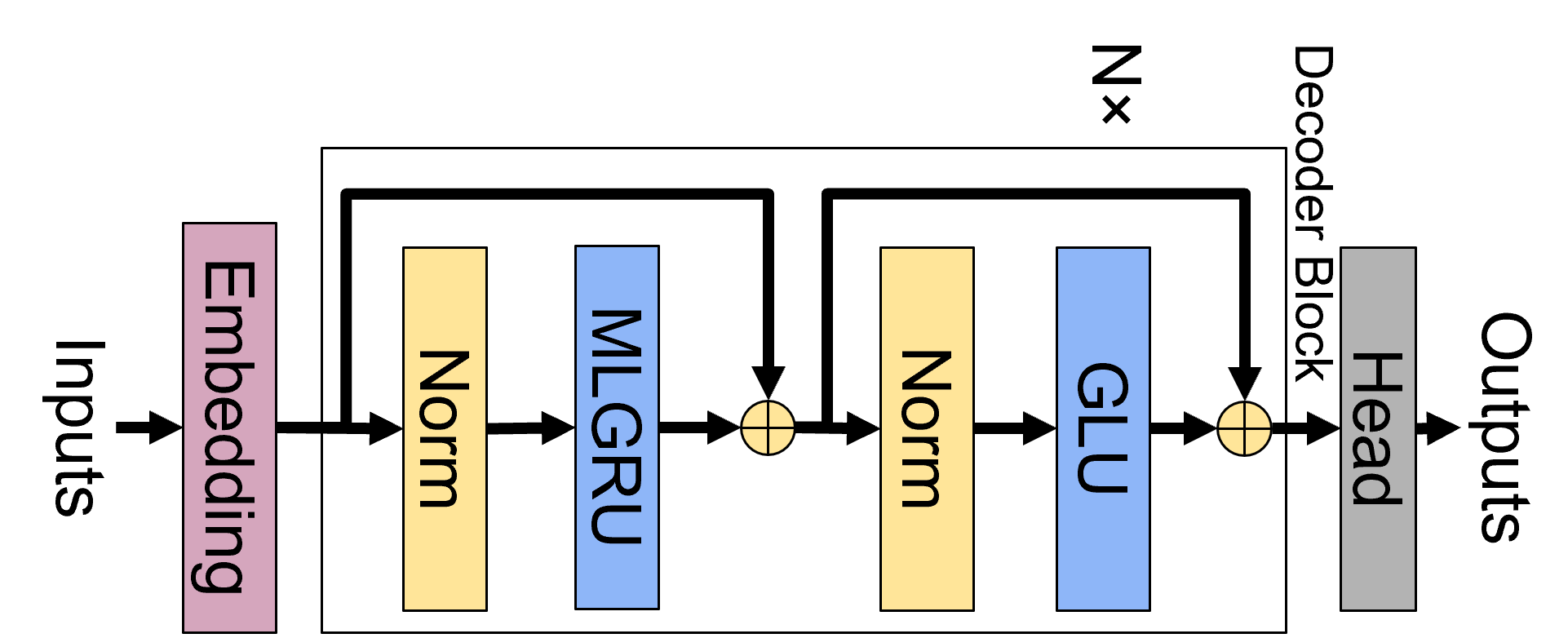}}
    \caption{MatMul-free LM}
    \label{MatMul-free LM}
\end{figure}

Fig. \ref{MatMul-free LM} shows the architecture of MatMul-free LM.
The input tokens are first embedded into vectors, just like in a standard transformer. These vectors are then fed into the decoder block, consisting of an MLGRU, a gated linear unit (GLU) \cite{dauphin2017language, shazeer2020glu}, root mean square (RMS) Norm \cite{RMSNorm}, and residual connections.
The decoder block is repeated N times (where N is the number of layers). Finally, the processed vectors are passed through a linear layer with ternary weights (called the heads) and then converted into tokens based on the logits.

\begin{figure}[!t]
    \centerline{\includegraphics[width=1.0\linewidth]{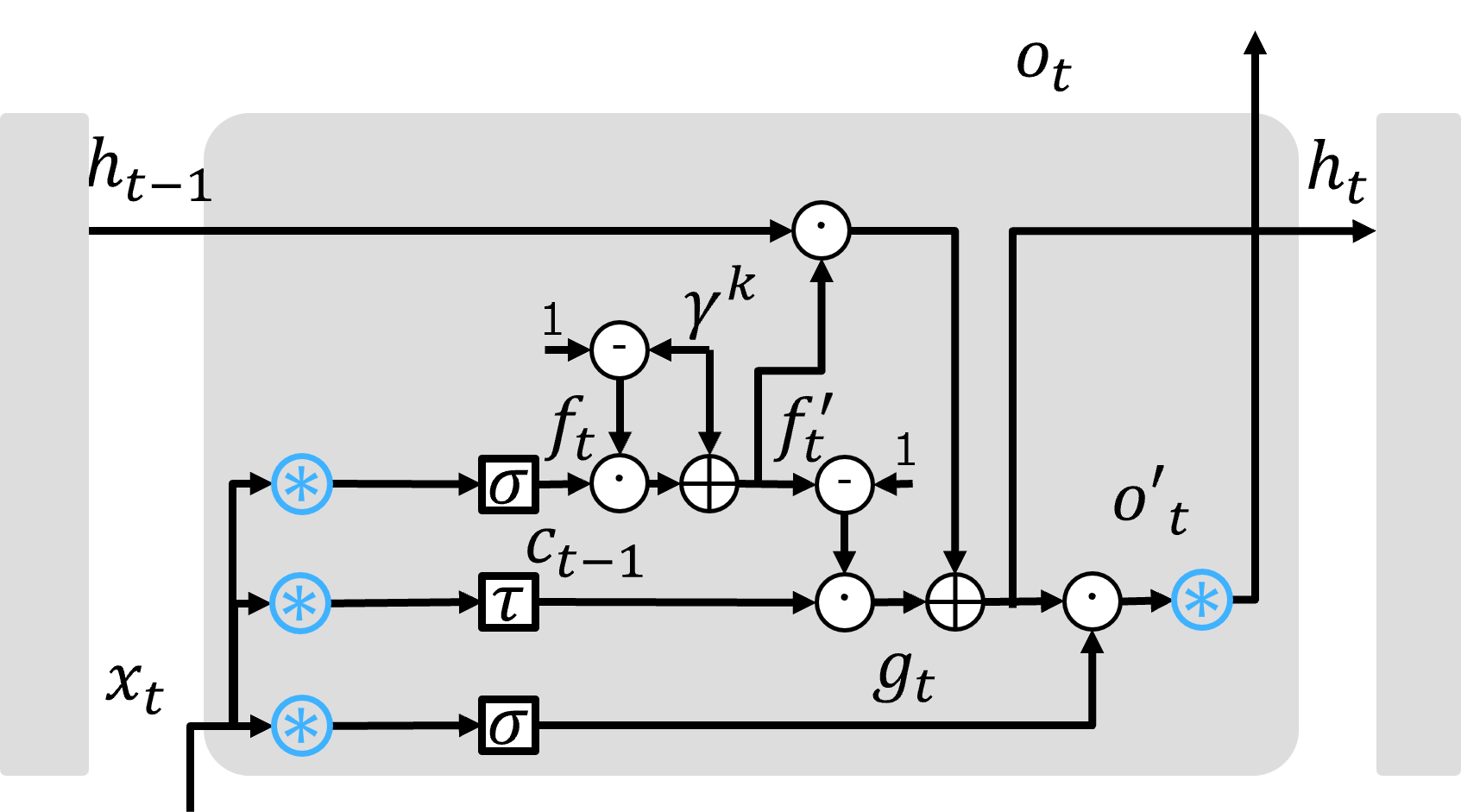}}
    \caption{Repeating module of MLGRU}
    \label{MLGRUmodule}
\end{figure}

Fig. \ref{MLGRUmodule} shows the repeating module of MLGRU in MatMul-free LM, and its computation is defined from \eqref{MLGRU} to \eqref{MLGRUend}.
\begin{align}
    \bm{f}_t &= \sigma \left( \bm{x}_t \circledast \bm{W}_f + \bm{b}_f \right) , 
    \label{MLGRU}\\
    \bm{f}'_t &= \bm{\gamma}^k + (\bm{1} - \bm{\gamma}^k) \odot \bm{f}_t, \label{lowerbound}\\
    \bm{c}_t &= \tau \left( \bm{x}_t \circledast \bm{W}_c + \bm{b}_c \right), \label{equ: c_t}\\ 
    \bm{h}_t &= \bm{f}'_t \odot \bm{h}_{t-1} + (\bm{1} - \bm{f}'_t) \odot \bm{c}_t ,\\
    \bm{g}_t &= \sigma \left( \bm{x}_t \circledast \bm{W}_g + \bm{b}_g \right) , \label{equ: g_t}\\
    \bm{o}'_t &= \bm{g}_t \odot \bm{h}_t,\\
    \bm{o}_t &= \bm{o}'_t \circledast \bm{W}_o + \bm{b}_o.\label{MLGRUend}
\end{align}
In Fig. \ref{MLGRUmodule}, the symbols $\odot$ and $\circledast$ represent element-wise multiplication and ternary matrix multiplication, respectively. Let the dimensionality of the vector be $d$, the index of the layer is $k$, the vector $\bm{x}_t \in \mathbb{R}^{d}$ is input from the previous component at time step $t$, and $\bm{h}_t \in \mathbb{R}^{d}$ is the internal state vector. The matrices $\bm{W}_c, \bm{W}_f, \bm{W}_o, \bm{W}_g \in \mathbb{R}^{d \times d}$ are ternary weight matrices, while $\bm{b}_c, \bm{b}_f, \bm{b}_o, \bm{b}_g \in \mathbb{R}^{d}$ are the corresponding biases terms. The vector $\bm{1}$ refers to a vector where all components are one. The vectors $\bm{c}_t, \bm{f}_t, \bm{f}'_t, \bm{\gamma}^k, \bm{g}_t, \bm{o'}_t, \bm{o}_t \in \mathbb{R}^{d}$ represent input vector, forget gate, updated forget gate, lower bound vector for the k-th layer, output gate, intermediate output, and final output, respectively. $\sigma$ denotes the sigmoid activation function, and $\tau$ denotes the  sigmoid-weighted linear unit (SiLU) activation function \cite{SiLU}. Compared to the traditional equations of long short-term memory (LSTM) or GRU, many dependencies on the previous hidden state $h_{t-1}$ are intentionally removed in this architecture to improve execution speed and computational efficiency. Nevertheless, the model still has a large number of parameters and requires considerable computational cost.

Equation \eqref{lowerbound} shows the calculation for adjusting $\bm{f}_t$ to $\bm{f}'_t$ so that it exceeds the lower bound. To obtain the variable $\bm{\gamma}^k$, the model employs the cumax activation function \cite{ON-LSTM} during training to calculate the forget gate values of the upper layers closer to 1, as in HGRN \cite{hgrn}. This cumax activation function, used only during training to calculate the lower bounds, is defined in \eqref{cummaxP} to \eqref{cumsum}.
\begin{align}
    \bm{P} &= \mathrm{Softmax}(\bm{\Gamma}, \mathrm{dim=0}), 
    \label{cummaxP}
    \\
    \bm{\gamma}^k = \mathrm{cumax}(\bm{\Gamma}) &= [\mathrm{Cumsum} (\bm{P}, \mathrm{dim=0})]_k, 
    \label{cummaxg} \\
    [\mathrm{Cumsum}(z)]_k &= (\Sigma^k_{i=1} z_i) - z_1.
    \label{cumsum}
\end{align}
The matrix $\bm{\Gamma} \in \mathbb{R}^{N \times d}$ is a trainable weight matrix used to compute $\bm{\gamma}^k$, where $N$ is the total number of layers. The intermediate output is represented by $\bm{P} \in \mathbb{R}^{N \times d}$. The cumulative sum (cumsum), as defined in \eqref{cumsum}, is used in this calculation. $\bm{z}$ denotes vectors that are used as inputs of cumsum. During inference, the model uses the $\bm{\gamma}$ values pre-computed and stored in memory, performing only the update process described in \eqref{lowerbound}.

\begin{figure}[!t]
    \centerline{\includegraphics[width=0.8\linewidth]{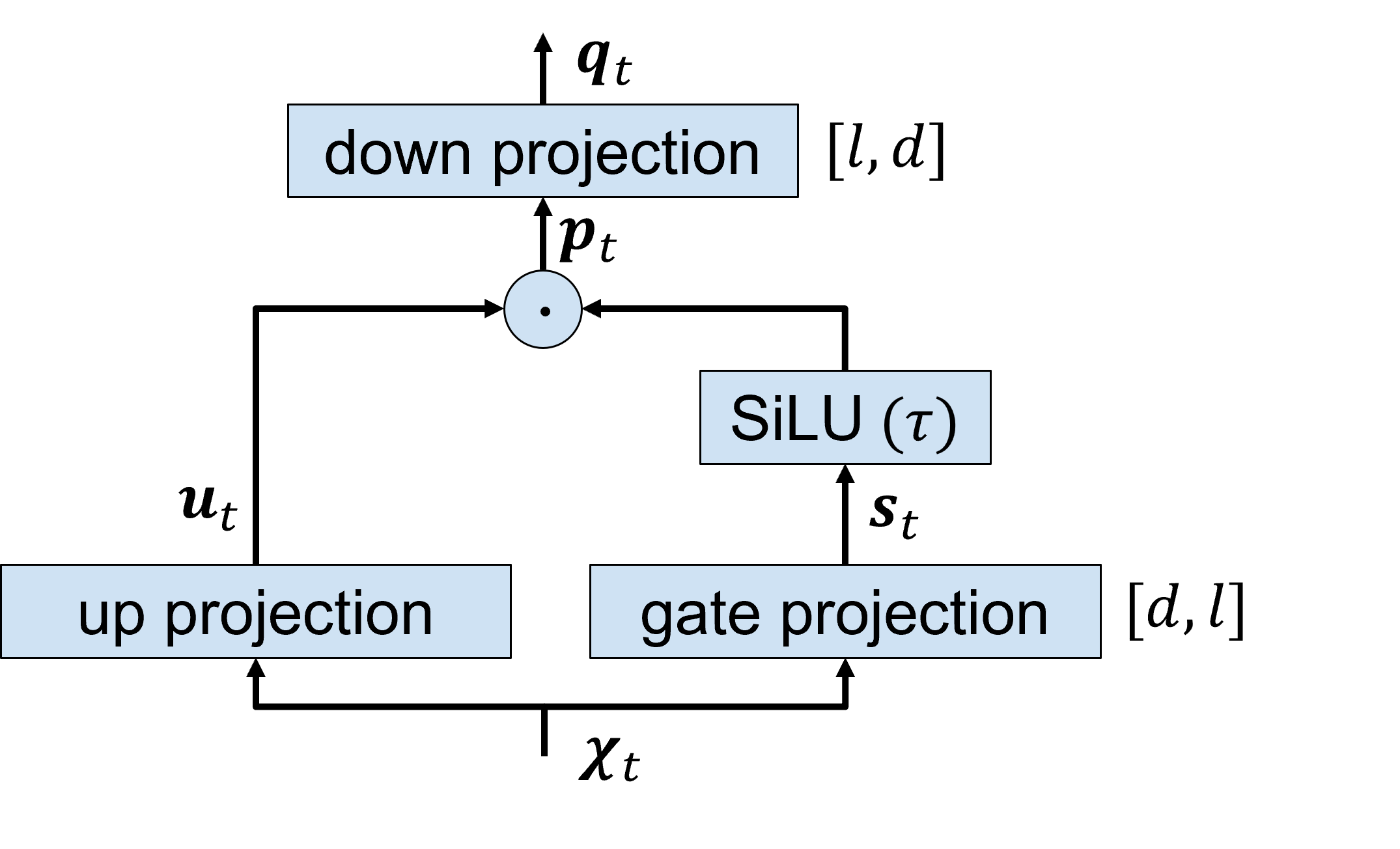}}
    \caption{GLU}
    \label{GLU_arch}
\end{figure}

Fig. \ref{GLU_arch} shows the GLU component, which is used as the FFN in this architecture, and \eqref{GLU} to \eqref{GLUend} define its computational process.
\begin{align}
    \bm{s}_t &=  \bm{\chi}_t \circledast \bm{W}_s,
    \label{GLU}\\
    \bm{u}_t &=  \bm{\chi}_t \circledast \bm{W}_u ,\label{equ:up_proj}\\
    \bm{p}_t &=  \tau(\bm{s}_t) \odot \bm{u}_t ,\\
    \bm{q}_t &=  \bm{p}_t \circledast \bm{W}_q ,
    \label{GLUend}
\end{align}
In the GLU, the following three-step process is performed:
\begin{enumerate}
\item Map inputs of the GLU, $\bm{\chi}_t \in \mathbb{R}^{d}$ to higher dimensions into $\bm{s}_t$ and $\bm{u}_t \in \mathbb{R}^{l}$ through the gate projection and up projection using $\bm{W}_s, \bm{W}_u \in \mathbb{R}^{d \times l}$, respectively.
\item Perform pointwise multiplication between $\bm{g}_t$, obtained by applying the SiLU activation function, and $\bm{u}_t$ applied to obtain representation $\bm{p}_t$.
\item Map $\bm{p}_t$ into $\bm{q}_t \in \mathbb{R}^{d}$ through a down projection using $\bm{W}_q \in \mathbb{R}^{d}$.
\end{enumerate}

\subsection{RC}
RC is a type of RNN that consists of only three layers: input layer, reservoir layer, and output (readout) layer. In contrast to deep learning, RC offers significantly lower training costs as it optimizes only the output layer parameters. Despite this simplicity, RC achieves competitive performance in various tasks.
Furthermore, incorporating the concept of the leaky integrator (LI) model enables the adjustment of the internal state dynamics of the reservoir to suit the task.

\begin{figure}[!t]
    \centerline{\includegraphics[width=1.0\linewidth]{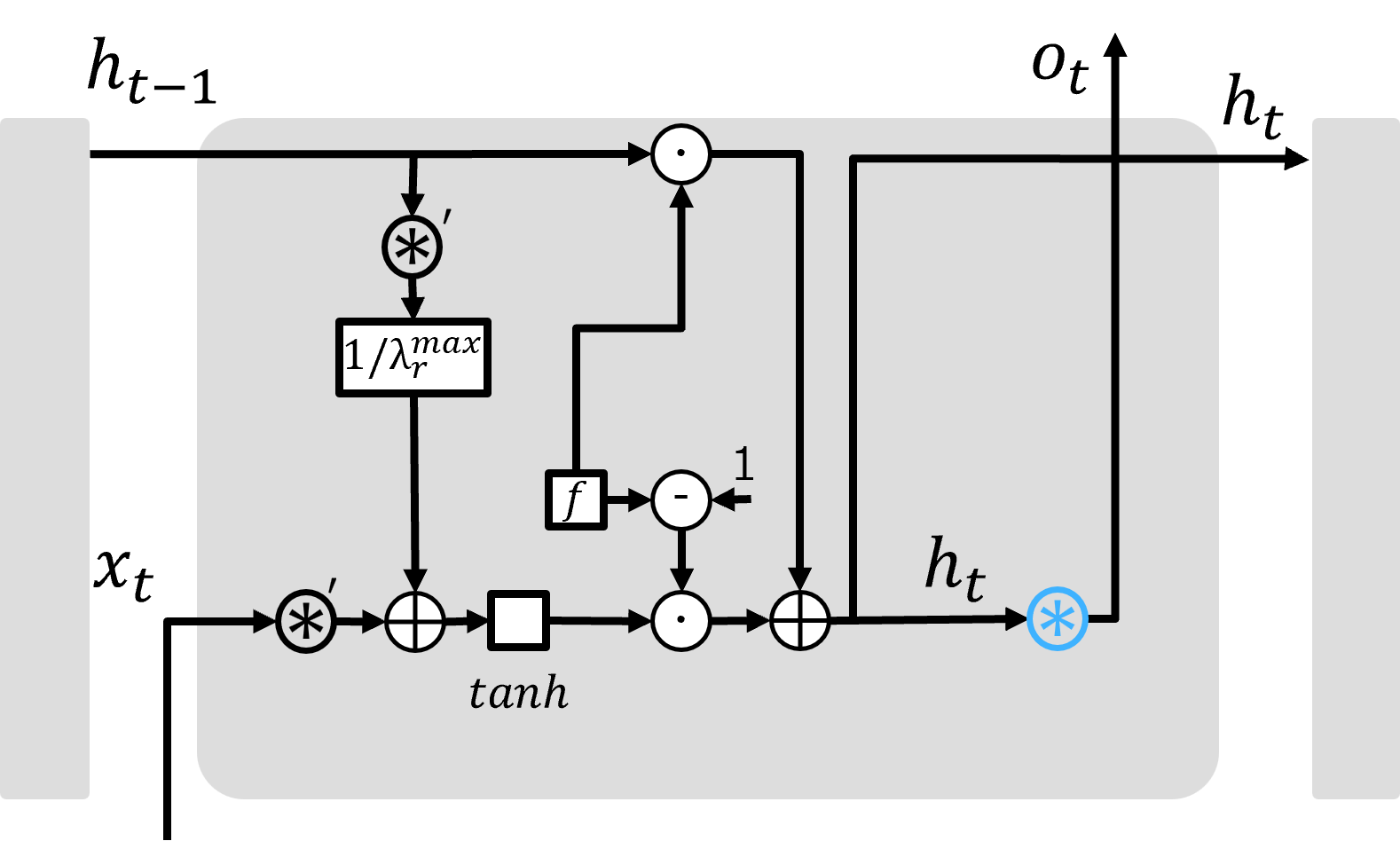}}
    \caption{Repeating module of ternary LI ESN}
    \label{RC_LIfig}
\end{figure}

Fig. \ref{RC_LIfig} and \eqref{ESN} to \eqref{ESNend} show a repeating module and processing of the ternary LI echo state network (ESN) \cite{ESN}, a type of RC, respectively.
In Fig. \ref{RC_LIfig}, ${\circledast}'$ denotes a ternary matrix multiplication with fixed weights, while $\circledast$ denotes a ternary matrix multiplication with trainable weights.
\begin{align}
    \bm{c}_t &= tanh \left( \bm{x}_t \circledast \bm{\overline{W}}_c + \bm{h}_{t-1} \circledast {\frac{\bm{\overline{W}}_r}{\lambda^{max}_{r}}} + \bm{b}_c \right), 
    \label{ESN}\\
    \bm{h}_t &= f\bm{h}_{t-1} + (1 - f) \bm{c}_t, \\
    \bm{o}_t &= \bm{h}_t \circledast \bm{W}_o + \bm{b}_o.
    \label{ESNend}
\end{align}
Here, $\bm{\overline{W}}_c$ is a ternary matrix with fixed weights, $\bm{\overline{W}}_r$ is a ternary sparse matrix with fixed weights, and $f$ represents the forget gate. ${\lambda^{max}_{r}}$ is the maximum eigenvalue of $\bm{\overline{W}}_r, \bm{c}_t, \bm{o}_t \in \mathbb{R}^{d}$ are input vector and output vector, respectively, and $\bm{h}_t \in \mathbb{R}^{l}$ is the internal state vector. $\bm{b}_c, \bm{b}_o \in \mathbb{R}^{d}$ are the biases. As shown from \eqref{ESN} to \eqref{ESNend}, the LI ESN is differentiable.

RC possesses echo state property, which ensures reproducibility in time-series signal processing. In other words, the reservoir dynamics exhibit a fading memory behavior, meaning that the input history uniquely determines the current state, while the influence of remote past inputs gradually decays.
This property is satisfied in ESNs with tanh activation if the spectral radius, the maximum absolute eigenvalue of $\bm{\overline{W}}_r$ is less than 1 \cite{ESN}.
Therefore, division by ${\lambda^{max}_{r}}$ is applied to scale the spectral radius \cite{ternalyESN}.

Due to the fading memory characteristics in reservoirs and the absence of gating mechanisms, a simple ESN often struggles to train long-term dependencies. To mitigate this limitation, the gated ESN \cite{GESN}, which introduces gating mechanism into the ESN, has been proposed. This model can be categorized into two types: one that trains only the output layer, and the other where the weights of the gate layer are trained by gradient descent. Both architectures exhibit superior performance relative to a simple ESN in natural language processing tasks including long-term dependencies.
\section{PROPOSED METHOD}
\subsection{RC MatMul-free LM}
\label{subsec: RC MM-free LM}
\begin{figure}[!t]
    \centerline{\includegraphics[width=1.0\linewidth]{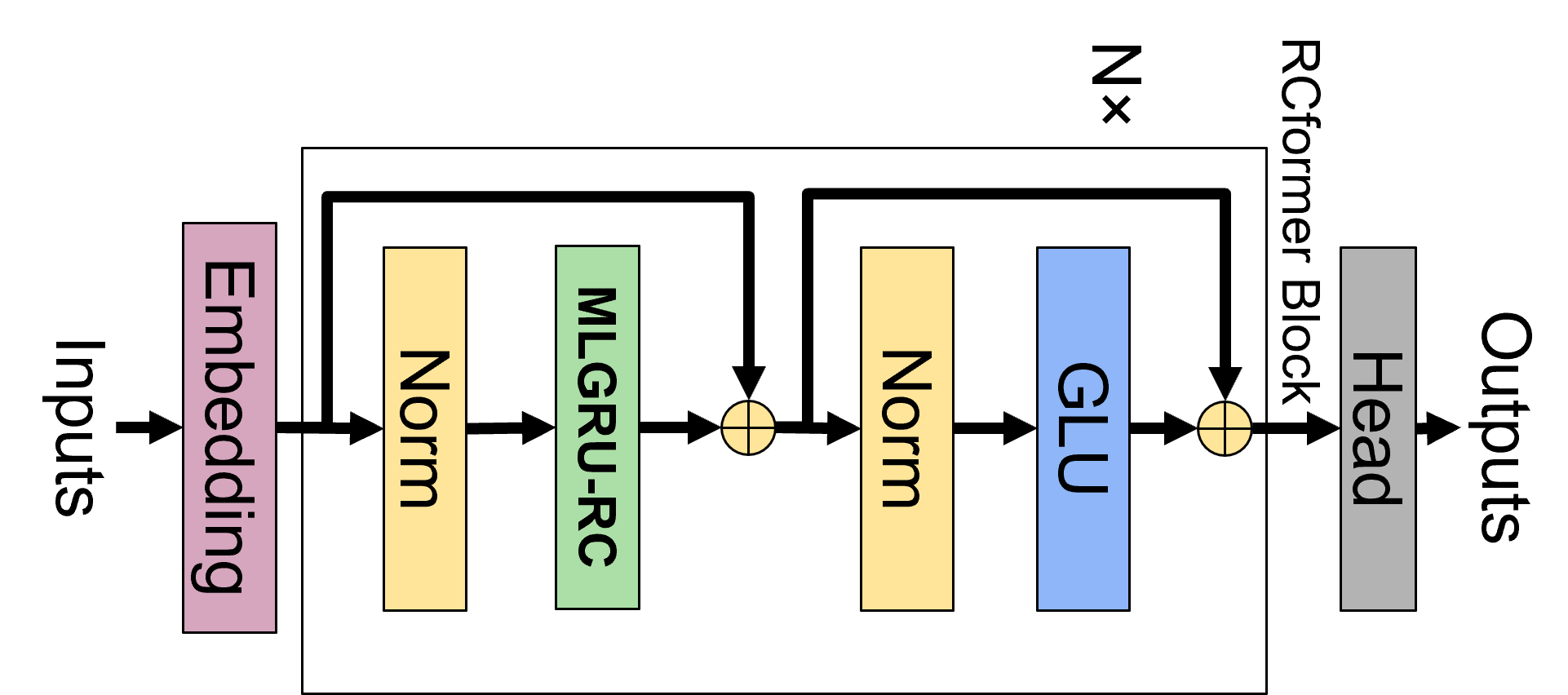}}
    \caption{RC MatMul-free LM}
    \label{RCMM-freeLM}
\end{figure}

\begin{figure}[!t]
    \centerline{\includegraphics[width=1.0\linewidth]{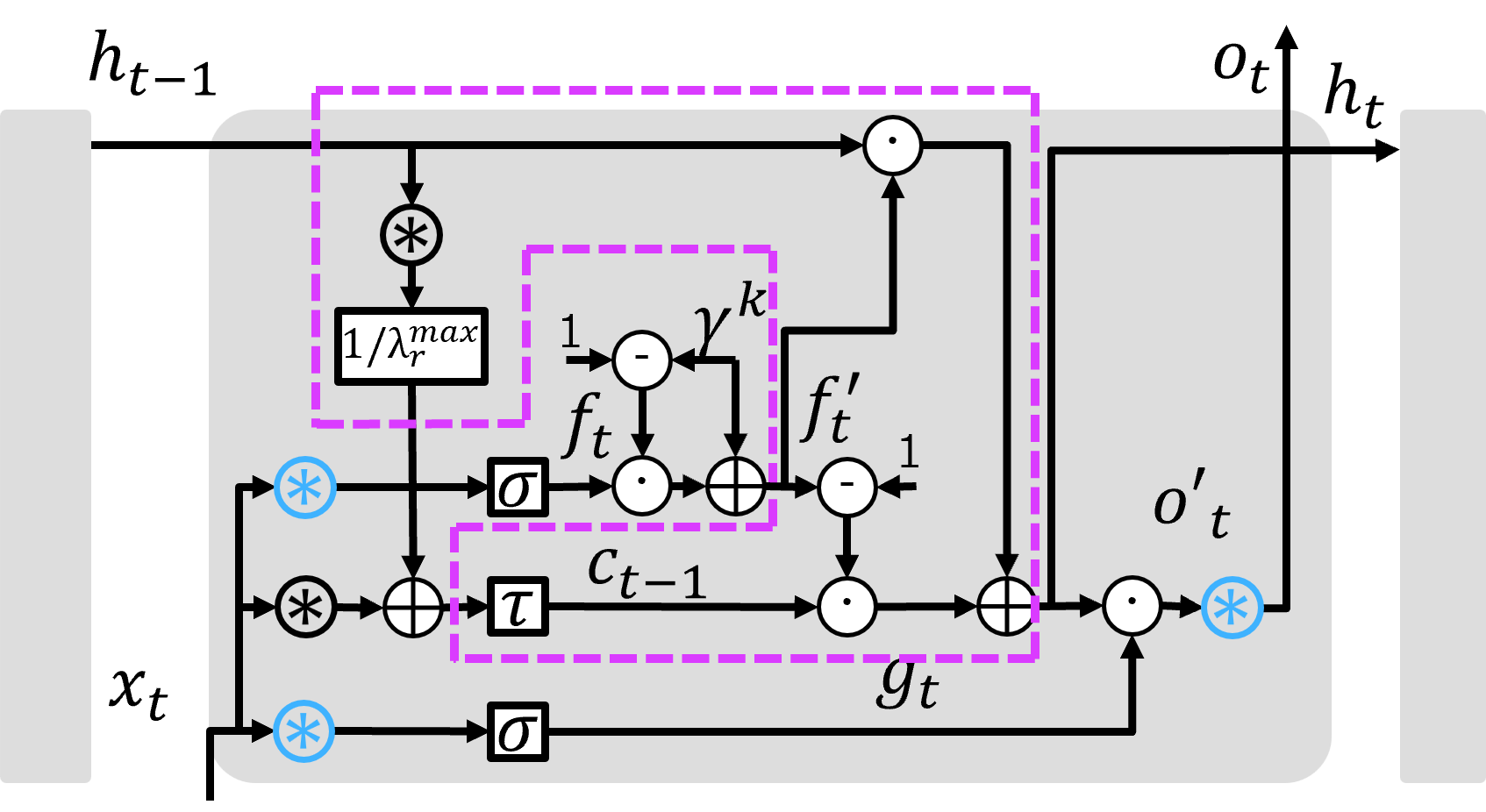}}
    \caption{Repeating module of MLGRU-RC}
    \label{MLGRU-RC}
\end{figure}

In this study, we propose RC MatMul-free LM, in which MLGRU of MatMul-free LM is replaced with MLGRU-RC inspired by RC. Fig. \ref{RCMM-freeLM} shows the architecture of this model.
As shown in Fig. \ref{RCMM-freeLM}, we name the decoder block of RC MatMul-free LM as the RCformer block. Fig. \ref{MLGRU-RC} shows the repeating module of MLGRU-RC.
The ternary matrix multiplication indicated by $\circledast '$ in Fig. \ref{MLGRU-RC} uses fixed random weights and shares the weights across all layers. This aims to reduce the number of parameters and computation time. In addition, this approach attempts to improve performance by adding a dependency on the internal state $\bm{h}_{t-1}$ to the processing of MatMul-free LM. $\circledast$ is a ternary logical matrix multiplication, which is updated via backpropagation similar to MatMul-free LM.

The proposed method modifies \eqref{equ: c_t} of the processing of MLGRU, shown in \eqref{MLGRU} to \eqref{MLGRUend}, according to \eqref{RC-inspired-MM-freeLLM}.
\begin{equation}
    \bm{c}_t = \tau \left( \bm{x}_t \circledast \bm{\overline{W}}_c + \bm{h}_{t-1} \circledast {\frac{\bm{\overline{W}}_r}{\lambda^{max}_{r}}} + \bm{b}_c \right),
        \label{RC-inspired-MM-freeLLM}
\end{equation}
Here, $\bm{\overline{W}}_r$ is a fixed sparse ternary weight matrix. Also, ${\lambda^{max}_{r}}$ is the maximum eigenvalue of $\bm{\overline{W}}_r$.
This approach omits the computation of the gradients $\frac{\partial L}{\partial \bm{\overline{W}}_c}$ and $\frac{\partial L}{\partial \bm{\overline{W}}_r}$ with respect to $\bm{\overline{W}}_c$ and $\bm{\overline{W}}_r$ when training this network, which lets $L$ as the loss function.
\subsection{Kernel Optimization of RC MatMul-free LM}
\begin{figure}[!t]
    \centerline{\includegraphics[width=1.0\linewidth]{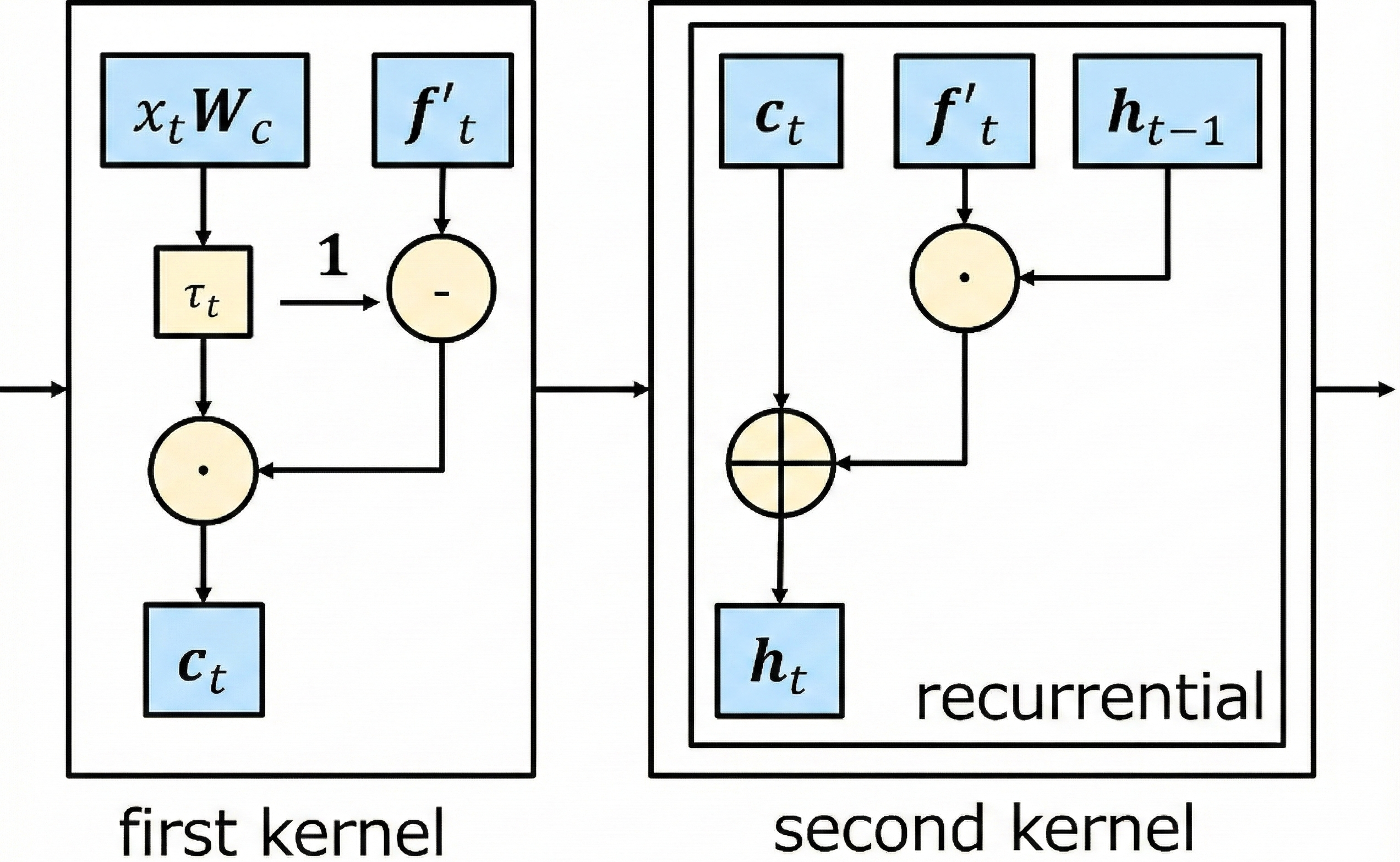}}
    \caption{Recurrent kernel of MatMul-free LM}
    \label{MatMul-freeLMkernel}
\end{figure}

\begin{figure}[!t]
    \centerline{\includegraphics[width=0.8\linewidth]{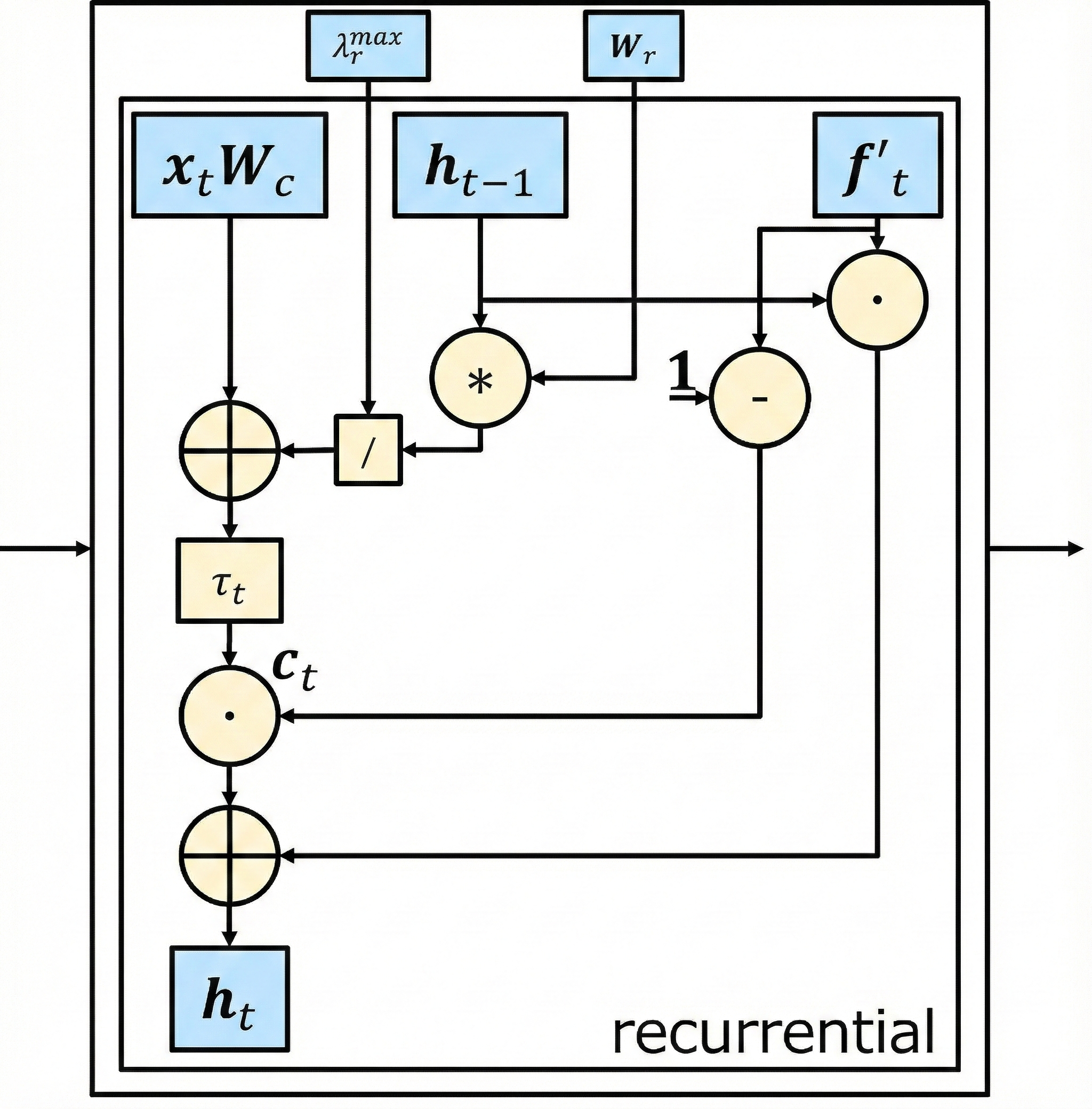}}
    \caption{Recurrent kernel of RC MatMul-free LM}
    \label{RC MatMul-freeLMkernel}
\end{figure}

RC MatMul-free LM reduced unnecessary memory read and write by fusing operations, which includes activation functions into the recurrent processing to achieve acceleration.
Fig. \ref{MatMul-freeLMkernel} shows a kernel of MatMul-free LM before improvement, which executes the operations corresponding to the area enclosed by the purple dashed line in Fig. \ref{MLGRU-RC}. Similarly, Fig. \ref{RC MatMul-freeLMkernel} shows a kernel of RC MatMul-free LM after improvement. Blocks highlighted in blue and outlined by black borders in Figs. \ref{MatMul-freeLMkernel} and \ref{RC MatMul-freeLMkernel} are parameter variables, and arrows connected to the blocks represent the read and write operations of the parameters. As shown in Fig. \ref{MatMul-freeLMkernel}, MatMul-free LM utilizes two kernels that cause extra reads and writes of $\bm{f}_t$ and $\bm{c}_t$. 
In RC MatMul-free LM, as shown in Fig. \ref{RC MatMul-freeLMkernel}, this approach integrated these operations into a single kernel. This integration reduces the redundancy of memory access and the overhead associated with kernel launches. 
Since the reads of $\lambda^{max}_r$ and $\bm{W}_r$ shown in Fig. \ref{RC MatMul-freeLMkernel} are not recurrent, their impact on processing time is almost negligible.

Furthermore, this improvement enables Triton\cite{triton} to optimize all operations shown in Fig. \ref{RC MatMul-freeLMkernel}. Consequently, this approach also reduces the processing time for matrix multiplication with $\bm{W}_r$ and calculations of an activation function.
\subsection{GRC MatMul-freeLM}
\label{Gfix MM-free LM}
To further reduce training time and the number of parameters, we propose a gated RC MatMul-free LM (GRC MatMul-free LM), which fixed $\bm{W}_f$ and $\bm{W}_g$ and shared them across all layers. In GRC MatMul-free LM, we extend \eqref{MLGRU} and \eqref{equ: g_t} of RC MatMul-free LM, which are formulated as \eqref{Fixed forget} and \eqref{Fixed goal}.
\begin{align}
    \bm{f}_t &= \sigma \left( \bm{x}_t \circledast \bm{\overline{W}_f} + \bm{b}_f \right), 
        \label{Fixed forget}\\
    \bm{g}_t &= \sigma \left( \bm{x}_t \circledast \bm{\overline{W}_g} + \bm{b}_g \right),
    \label{Fixed goal}
\end{align}
Similar to $\bm{\overline{W}}_c$, fixed ternary weight matrices $\bm{\overline{W}}_f$ and $\bm{\overline{W}}_g$ are initialized with random values; consequently, computing their corresponding gradients $\frac{\partial L}{\partial \bm{\overline{W}}_f}$ and $\frac{\partial L}{\partial \bm{\overline{W}}_g}$ is not required. 

\begin{figure}[!t]
    \centerline{\includegraphics[width=1.0\linewidth]{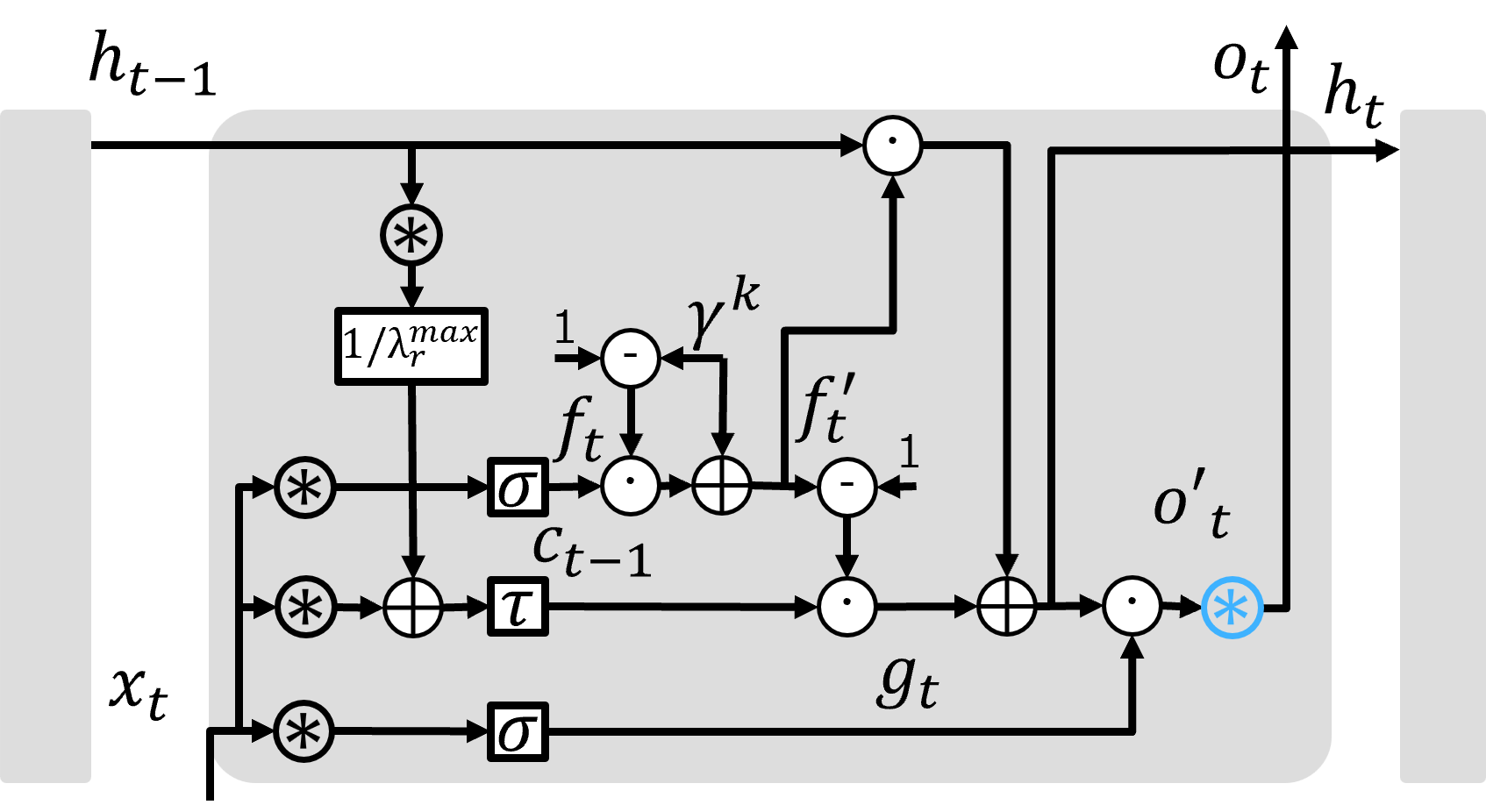}}
    \caption{Repeating module of gated MLGRU-RC}
    \label{GfixMLGRU-RC}
\end{figure}

Although fixing weights in RNNs with only the same-dimensional mappings generally tends to degrade performance, we consider that fixing $\bm{W}_f$ retains its function as a forget gate. That is because $\bm{f}'_t$ monotonically increases as the network becomes deeper, resulting in the shallow and deep layers capturing short-term and long-term dependencies, respectively. Furthermore, $\bm{f}'_t$ is tuned coarsely via $\bm{\gamma}^k$ that is computed from trainable $\bm{\Gamma}$. Therefore, we adopt this extension.
Regarding $\bm{g}_t$, we also apply this extension. This is inspired by the observation that fixing and sharing all attention mechanism parameters makes minimal impact on performance \cite{albert}. Fig. \ref{GfixMLGRU-RC} shows the repeating module of MLGRU-RC in GRC MatMul-freeLM.
As with RC MatMul-free LM, we also optimize the kernel in GRC MatMul-free LM.
\section {EXPERIMENT}
In this experiment, we developed training programs for RC MatMul-free LM and GRC MatMul-free LM using PyTorch\cite{pytorch} and Triton\cite{triton}. Table \ref{ex_setting} summarizes the experimental settings. In Table \ref{ex_setting}, sparsity denotes the sparsity rate of $\bm{\overline{W}}_r$, and context size shows the maximum number of tokens the model can process at once. The reason for using a smaller context size compared to that in previous work \cite{MM-freeLM} is to compare model performance attributable to the architecture within a shorter training time. SlimPajama dataset shown in Table \ref{ex_setting} is a large-scale natural language dataset containing 627B tokens. For this experiment, we used randomly sampled data from the 627B tokens. Regarding random seeds for CUDA, we used the same value across all experiments.

For the benchmark tasks, we adopted ARC Easy (ARCe), ARC Challenge (ARCc) \cite{clark2018thinksolvedquestionanswering}, HellaSwag (Hs) \cite{zellers-etal-2019-hellaswag}, Openbook QA (OQ) \cite{openbookqa}, PIQA (PQ) \cite{piqa}, and Winogrande (WGe) \cite{winogrande}. We evaluated these tasks using LM Evaluation Harness\cite{eval-harness}, a commonly used framework for consistent evaluation of multiple benchmark tasks in LLM performance assessment.

\begin{table*}[!t]
    \centering
    \caption{Setting of experiments}
    \label{ex_setting}
    \begin{tabular}{lcccccc}
    \toprule
    \toprule
    \textbf{Dataset} & \textbf{Tokenizer} & \textbf{\makecell{Number of Token \\ (Train / Eval)}} & \textbf{Machine}  & \textbf{Sparsity of reservoir layer} & \textbf{Context Size}\\
    \midrule
    SlimPajama\cite{cerebras2023slimpajama} & Mistral & 15 B/541 M & One H100 GPU & 85\% & 128\\
    \bottomrule
\end{tabular}
\end{table*}

\subsection{Learning Rate}
Since MatMul-free LM \cite{MM-freeLM} reported that $0.01$ is the best learning rate, this study explored parameters around this value and adopted $\frac{0.01}{\sqrt{8}}$ as the learning rate at which training converged with batch size 256 in this experiment. We used a cosine scheduler as the learning rate scheduler.

\subsection{Result}
In this experiment, we evaluated two types of weights from the xavier uniform \cite{xavier_uniform} distribution and averaged the results across these runs. Tables \ref{speed}, \ref{loss}, and \ref{bench score} show the details of the models and computation time, the loss of models, and the benchmark scores with their corresponding averages, respectively. In Table \ref{speed}, the values in parentheses under size indicate the number of fixed parameters that are not updated during training. For memory usage of parameters, we calculated the memory usage required for parameters during inference. We computed the memory size of ternary parameters as $\log_2{3}$-bit.

\begin{table*}[!t]
    \centering
    \caption{Size and runtime of each model}
    \label{speed}
    \begin{tabular}{lccccc}
    \toprule
    \toprule
    \textbf{Models} & \textbf{\makecell{Total size [M] \\ (Size of fixed [M])}} & \textbf{\makecell{Memory usage of\\ parameters [MB]}} & \textbf{\makecell{Train\\runtime [h]}} & \textbf{\makecell{Eval\\runtime [min]}}\\
    \midrule
    MatMul-free LM \cite{MM-freeLM} & 374 & 127 & 73.61 & 43.68\\
    RC MatMul-free LM (\ref{subsec: RC MM-free LM}) & 351(2) & 122 & 70.77 & 41.00\\
    GRC MatMul-free LM (\ref{Gfix MM-free LM}) & $\bm{303(4)}$ & $\bm{113}$ & $\bm{66.32}$ & $\bm{40.18}$\\
    \bottomrule
\end{tabular}
\end{table*}
\begin{table*}[!t]
    \centering
    \begin{threeparttable}
    \caption{Loss of each model}
    \label{loss}
    \begin{tabular}{lccccc}
    \toprule
    \toprule
    \textbf{Models} & \textbf{Train loss} & \textbf{Eval loss}\\
    \midrule
    MatMul-free LM \cite{MM-freeLM} & $\bm{3.291}$ & $\bm{2.995}$\\
    RC MatMul-free LM (\ref{subsec: RC MM-free LM}) & 3.349 & 3.048\\
    GRC MatMul-free LM (\ref{Gfix MM-free LM}) & 3.476 & 3.153\\
    \bottomrule
\end{tabular}
\end{threeparttable}
\end{table*}
\begin{table*}[!t]
    \centering
    \begin{threeparttable}
    \caption{Benchmark score of each model}
    \label{bench score}
    \begin{tabular}{lccccccccc}
    \toprule
    \toprule
    \textbf{Models} & \textbf{ARCc} & \textbf{ARCe} & \textbf{Hs} & \textbf{OQ} & \textbf{PQ} & \textbf{WGe} & \textbf{Avg.}\\
    \midrule
    MatMul-free LM \cite{MM-freeLM} & $\bm{23.8}$ & $\bm{42.4}$ & $\bm{34.0}$ & $\bm{29.6}$ & $\bm{63.4}$ & 48.8 & $\bm{40.3}$\\
    RC MatMul-free LM (\ref{subsec: RC MM-free LM}) & 23.3 & 42.1 & 32.2 & 27.3 & 62.0 & 49.7 & 39.4\\
    GRC MatMul-free LM (\ref{Gfix MM-free LM}) & 23.5 & 41.5 & 30.6 & 29.0 & 61.2 & $\bm{50.7}$ & 39.4\\
    \bottomrule
\end{tabular}
\end{threeparttable}
\end{table*}

Table \ref{speed} confirms that improving MatMul-free LM to RC MatMul-free LM or GRC MatMul-free LM reduces the computation time for both training and inference. Specifically, for training, RC MatMul-free LM reduces the time to 
3.9\%, and GRC MatMul-free LM reduces it to 9.9\%. For inference, RC MatMul-free LM and GRC MatMul-free LM reduce the time to 6.1\% and 8.0\%, respectively. Regarding memory usage, 
62.5 MB out of the total memory consumption shown in the table corresponds to the memory usage for the embedding layer.
Meanwhile, Table \ref{loss} and \ref{bench score} show that the performance differences between MatMul-free LM, RC MatMul-free LM, and GRC MatMul-free LM are minimal, confirming that the improvements maintain model performance.
\section {DISCUSSION}
Table \ref{speed} shows a reduction in model size, training time, and inference time. We attribute this decrease to parameter fixing, parameter sharing, and kernel optimizations, all of which reduce the number of backpropagation operations and memory accesses. Although the ratio appears small, scaling RC MatMul-free LM to a size comparable to the practical Llama3 405 B would translate that reduction into a saving of 2 to 5 days of training time.

Conversely, Tables \ref{loss} and \ref{bench score} show that MatMul-free LM suffers almost no performance degradation even when fixing and sharing its parameters, similar to the full-bit transformer. This result implies that RNN-based LLM or any ternary LLM will likewise lose little accuracy under parameter fixing or sharing, and that even a token-mixer component inspired by RC could remain functional.

Because this study confirms the effectiveness of parameter fixing for QAT of ternary LLMs, we expect to shorten training time further by reducing memory accesses through the same kernel optimizations used at inference time \cite{optq}. Specifically, the frozen linear functions use $\log_2 3$-bit rather than 16-bit weights; by packing weights and reading or writing them together, we can reduce the number of memory accesses to approximately $\frac{1}{10}$ times that of the 16-bit case. We can achieve this improvement by adapting the ternary optimized linear functions already employed during inference.

We confirmed that the network can still train when we use a component inspired by RC as the token-mixer. Future work will therefore explore incorporating existing acceleration and performance enhancement techniques for RC into RC MatMul-free LM.

The layers whose parameters we froze act as random, same-dimensional mappings, so there remains room for further performance improvement. We need to consider introducing higher-dimensional mappings. However, a naive introduction increases the number of parameters in the unshared output layers and hence the total model size. So we must devise alternatives for those layers. 

Although we did not verify it experimentally, the fact that the only changes are parameter fixing and the addition of new layers—while maintaining comparable performance—suggests that RC MatMul-free LM, like MatMul-free LM, should obey the usual scaling laws.

We trained RC MatMul-free LM and performed inference on GPUs using PyTorch \cite{pytorch} and Triton \cite{triton}. Because RC lends itself to physical implementation, and RC MatMul-free LM was developed with field programmable gate array deployment in mind, we likewise expect physical or circuit implementations of RC MatMul-free LM.
\section {CONCLUSION}
In this study, we propose RC MatMul-free LM, which introduces the concept of RC into MatMul-free LM to reduce the number of parameters and memory usage, one of the main causes of the high computational cost that hinders the practical deployment of large language models.
In the proposed RC MatMul-free LM, we replace MLGRU in the previous MatMul-free LM with MLGRU-RC, an RC-based component, thereby achieving inter-layer parameter fixing and sharing.

Experiments on the large-scale natural-language dataset SlimPajama demonstrate that, compared with the previous MatMul-free LM, the proposed RC MatMul-free LM reaches comparable performance while reducing the parameter count by up to 19\% and shortening training and inference times by 9.9\% and 8.0\%, respectively.

These results constitute a significant step toward resolving the computational cost problem of LLMs.
\section*{Acknowledgments}
This research is based on results obtained from a project,
JPNP16007, commissioned by the New Energy and Industrial Technology Development Organization (NEDO). This
work received support from JSPS KAKENHI Grant Number 23H03468, as well as from JST ALCANext Grant Number JPMJAN23F3.
\renewcommand{\refname}{\MakeUppercase{References}}
\bibliographystyle{myieeetr}
\bibliography{citation}

\vspace{12pt}

\end{document}